%% file: acl2020.tex
\documentclass[11pt,a4paper]{article}
\usepackage[hyperref]{emnlp2020}
\usepackage{times}
\usepackage{latexsym}

\usepackage{amsthm,booktabs}
\usepackage{amsmath}

\usepackage{todonotes}
\usepackage[margin=0.5cm]{subcaption}

\usepackage{placeins}
\usepackage{paralist}

\theoremstyle{definition}

\usepackage{microtype}

\aclfinalcopy 

\input{math_commands.tex}

\title{Emergent Language Generalization and Acquisition Speed are not tied to Compositionality}

\author{Eugene Kharitonov \\
  Facebook AI \\
    \texttt{kharitinov@fb.com} \\
\And
Marco Baroni \\
  Facebook AI, ICREA \\
  \texttt{mbaroni@fb.com} 
  \\}

\date{}

\begin{document}
\maketitle
\begin{abstract}
Studies of discrete languages emerging when neural agents communicate to solve a joint task often look for evidence of compositional structure. This stems for the expectation that such a structure would allow languages to be acquired faster by the agents and enable them to generalize better. We argue that these beneficial properties are only loosely connected to compositionality. In two experiments, we demonstrate that, depending on the task, non-compositional languages might show equal, or better,  generalization performance and acquisition speed than compositional ones. Further research in the area should be clearer about what benefits are expected from compositionality, and how the latter would lead to them.
\end{abstract}

\section{Introduction}

There is a recent spike of interest in studying the languages that emerge when artificial neural agents communicate to solve a common task \cite{Foerster:etal:2016,Lazaridou2016,Havrylov2017}. A good portion of such studies looks for traces of \textit{compositional} structure in those languages, or even tries to inject such structure into them \cite{Kottur:etal:2017,Choi:etal:2018,Lazaridou:etal:2018,Mordatch:Abbeel:2018,Andreas2019,cogswell2019,Li2019,Resnick2019,Chaabouni2020}. Besides possibly providing insights on how compositionality emerged in natural language \cite{Townsend:etal:2018}, this emphasis is justified by the idea that a compositional language has various \emph{desirable} properties. In particular, compositional languages are expected to help agents to better \emph{generalize} to new (composite) inputs \cite{Kottur:etal:2017,Lazaridou:etal:2018}, and to be faster to acquire~\cite{cogswell2019,Li2019,Ren:etal:2019}. 


We engage here with this ongoing research pursuit. We step back and reflect on the benefits that compositionality can bring to the emergent languages: if there is none, then it is unlikely that agents will develop compositional languages on their own. Indeed, several studies have shown that compositionality does not emerge naturally among neural agents \cite[e.g.][]{Kottur:etal:2017,Lazaridou:etal:2018,Andreas2019}. On the other hand, understanding what benefits compositionality could bring to a language would help us in establishing the conditions for its emergence. 


Compositionality is typically seen as a property of a language, independent of the task being considered. However, the task will likely influence properties such as generalization and ease of acquisition, that compositionality is expected to correlate with. Our experiments show that it is easy to construct tasks for which a compositional language is equally hard, or harder, to acquire and does not generalize better than a non-compositional one. %
Hence, language emergence researchers need to be clear about \begin{inparaenum}[i)]
    \item which benefits they expect from compositionality and
    \item in which way compositionality would lead to those benefits in their setups.
\end{inparaenum} 
Otherwise, the agents will likely develop perfectly adequate communication systems that are not compositional.

\section{Operationalizing compositionality}
\label{sReview}
Before we proceed, let us clarify our definition of compositionality. Linguists and philosophers have extensively studied the topic for centuries \cite[see][for a thorough review]{Pagin:Westerstahl:2010b,Pagin:Westerstahl:2010}. However, the standard definition that a language is compositional if the meaning of each expression in it is a function of the meaning of its parts and the rules to combine them is so general as to be vacuous for our purposes (under such definition, even the highly opaque languages we will introduce below are compositional, \emph{contra} our intuitions). 

In most current language emergence research, the input to language is composite in the sense that it consists of ensembles of elements. In this context, intuitively, a language is compositional if its symbols denote input elements in a \emph{disentangled} way, so that they can be freely juxtaposed to refer to arbitrary combinations of them. More precisely, the following property might suffice for a limited but practical characterization of compositionality. Given a set of atomic input elements (for example, a set of independent attribute values), each atomic symbol should refer to one and only one input element, \emph{independently of the other symbols it co-occurs with}.\footnote{We leave the definition of what counts as an atomic symbol open: it could be a single character, a character bound to a certain position in a message string, a character sequence, etc.} A language where all symbols respect this property is compositional in the intuitive sense that, if we know the symbols that denote a set of input elements, we can assemble them (possibly, following some syntactic rules of the language) to refer to the ensemble of those input elements, irrespective of whether we have ever observed the relevant ensemble. Consider for example a world where inputs consist of two attributes, each taking a number of values. A language licensing only two-character sequences, where the character in the first position refers to the value of the first attribute, and that in the second position independently refers to the value of the second, would be compositional in our sense. On the other hand, a language that also licenses two-character sequences, but where both characters in a sequence are needed to decode the values of both the first and the second input attribute, would not be compositional. We will refer to the lack of symbol interdependence in denoting distinct input elements as \emph{na\"{i}ve compositionality}.\footnote{Na\"{i}ve in the sense that it is only appropriate when complex meanings are ensembles of atomic meanings. The definition breaks down when complex meanings result from functions that merge their components in different ways than simple ensembling, as is often the case in natural language.} 

We believe that na\"{i}ve compositionality captures the intuition behind explicit and implicit definitions of compositionality in emergent language research. 
For example, \citet{Kottur:etal:2017} deem non-compositional those languages that either use single symbols to refer to ensembles of input elements, or where the meaning of a symbol depends on the context in which it occurs. 
\citet{Havrylov2017} looked for symbol-position combinations that encode a single concept in an image, as a sign of a compositional behavior. A na\"{i}vely compositional language will maximize the two recently proposed compositionality measures of residual entropy~\citep{Resnick2019} and positional disentanglement~\citep{Chaabouni2020}.







Na\"{i}ve compositionality is also closely related to the notion of \emph{disentanglement} in representation learning \cite{Bengio:etal:2013}. 
Interestingly, \citet{locatello2018challenging} reported that disentanglement is not necessarily helpful for sample efficiency in downstream tasks, as had been previously argued. This resonates with our results below.

\section{Communication Game}
We base our experimental study on a one-episode one-direction communication game, as commonly done in the relevant literature~\cite{ Lazaridou2016,Lazaridou:etal:2018,Havrylov2017,Chaabouni2019}. In this setup, we have two agents, Sender and Receiver. An input $\vi$ is fed into Sender, in turn Sender produces a message $\vm$, which is consumed by Receiver. Receiver produces its output $\hat \vo$. Comparing the output $\hat \vo$ with the ground-truth output $\vo$ provides a loss. We used EGG~\cite{Kharitonov2019} to implement the experiments.\footnote{The code is available at \url{https://github.com/facebookresearch/EGG/tree/master/egg/zoo/compositional_efficiency}.}

In contrast to the language emergence scenario, we use  a hard-coded Sender agent that produces a fixed, pre-defined language. This allows us to easily control the (na\"ive) compositionality of the language and measure how it affects Receiver's performance. This setup is akin to the motivating example of~\citet{Li2019}.

We study two Receiver's characteristics: (i) acquisition speed, measured as the number of epoch needed to achieve a fixed level of performance on training set, and (ii) generalization performance on held-out data. 

\begin{table}[t]
\begin{center}
\resizebox{\columnwidth}{!}{%
\begin{tabular}{lcccccccccr}
    \toprule
    \multicolumn{7}{c}{{\textbf{Acquisition speed}}} \\
    & \multicolumn{2}{c}{{task-identity}}  & \multicolumn{2}{c}{{task-linear}} & \multicolumn{2}{c}{{task-entangled}} \\
     \midrule
    \multicolumn{7}{c}{{LSTM}} \\
    lang-identity && $5.3_{\pm 0.1}$ && $30.0_{\pm 1.4}$ && $20.1_{\pm 0.6}$ \\
    lang-entangled && $20.1_{\pm 0.5}$ && $26.6_{\pm 0.8}$ &&  $5.5_{\pm 0.2}$ \\
    \midrule
    \multicolumn{7}{c}{{GRU}} \\
    lang-identity && $5.6_{\pm 0.2} $&& $94.0_{\pm 7.7}$ && $57.4_{\pm 13.9}$ \\
    lang-entangled && $37.2_{\pm 2.7}$ && $91.5_{\pm 4.7}$ &&  $5.7_{\pm0.2}$ \\

     \midrule
    \multicolumn{7}{c}{{\textbf{Test accuracy}}} \\
    & \multicolumn{2}{c}{{task-identity}}  & \multicolumn{2}{c}{{task-linear}} & \multicolumn{2}{c}{{task-entangled}} \\
     \midrule
    \multicolumn{7}{c}{{LSTM}} \\
    lang-identity && $0.97_{\pm 0.00}$ && $0.0_{\pm 0.0}$ && $0.06_{\pm 0.01}$ \\
    lang-entangled && $0.08_{\pm 0.01}$ && $0.0_{\pm 0.0}$ &&  $0.97_{\pm 0.00}$ \\
    \midrule
    \multicolumn{7}{c}{{GRU}} \\
    lang-identity && $0.97_{\pm 0.00} $&& $0.0_{\pm 0.0}$ && $0.06_{\pm 0.01}$ \\
    lang-entangled && $0.10_{\pm 0.02}$ && $0.0_{\pm 0.0}$ &&  $0.97_{\pm0.00}$ \\
    \bottomrule
    \end{tabular}
}
\caption{Attval experiment. Top: epochs to achieve perfect accuracy on training set. Bottom: test accuracy after convergence. $\pm$ marks 1 standard error of the mean.\vspace{-3mm}}
    \label{tab:attrval}
\end{center}
\end{table}

\section{Experimental setup}
To demonstrate that compositionality of a language alone, detached from the task at hand, does not necessarily lead to higher generalization or faster acquisition speed, we design two experiments. 

The first experiment (\textit{attval}) operates in an attribute-value world, similar to those of~\citet{Kottur:etal:2017,Chaabouni2019}. We fix two languages, one compositional and one not, and build three tasks: (i) ``easy'' for compositional language and ``hard'' for non-compositional; (ii) equally ``hard'' for both; (iii) ``hard'' for compositional language and ``easy'' for non-compositional language. Informally, we control the amount of computation needed by Listener to perform a task starting from a language, where it can be equally hard to rely on compositional or non-compositional languages, or the answers could even be readily available in a non-compositional language.

In the second experiment (\emph{coordinates}), we design a single task that is equally ``easy'' for an entire family of languages (parameterized by a continuous value), including compositional and non-compositional ones. The task is to transmit points on the 2D plane (thus, the input ensembles here are pairs of point coordinates). Here, we leverage the observation that a typical neural model has a linear output layer, for which it is equally easy to learn any rotation of the ground-truth outputs. Such rotation-group-invariance could play role in games where continuous image embeddings are used as input~\cite{Lazaridou2016, Havrylov2017}.

\subsection{Attval experiment}
\label{sTask}

\paragraph{Input} Sender's input $\vi$ is a two-dimensional vector;  each dimension encodes one of two attributes, each having $n_v$ values: $\vi \in \{1..n_v\} \times \{1..n_v \}$.
\paragraph{Languages} We consider two languages, with messages of length two and vocabulary size $n_v$. The first language, \textit{lang-identity}, represents the inputs as-is, by putting the value of the first (second) attribute in the first (second) position: $(m_1, m_2) \gets (i_1, i_2)$. In the second language, \textit{lang-entangled}, the first and the second positions are obtained as follows:
\begin{equation}
m_j \gets (i_1 + (-1)^j \cdot i_2) \mod n_v, \quad j \in \{1, 2\}
\label{eq:rotate}
\end{equation}

\textit{Lang-identity} and \textit{lang-entangled} have exactly the same $n_v^2$ utterances. While \textit{lang-identity} is na\"ively compositional (one symbol encodes one attribute only), \textit{lang-entangled} is not: each symbol of an utterance encodes equal amount of information about both attributes and both symbols are equally needed for decoding each attribute.

\paragraph{Tasks}
We consider three tasks. In all of them, Receiver outputs two discrete values, $\vo \in \{1..n_v\} \times \{1..n_v\}$.  In \textit{task-identity}, Receiver has to recover the original input of Sender, $\vi$.  In the second task, \textit{task-linear}, Receiver needs to output two values that are obtained as integer linear-modulo operations of the original input values: $\vo \gets A \cdot \vi + \vb \mod n_v$. In the third task, \textit{task-entangled}, we require Receiver to output $o_j \gets (i_1 + (-1)^j \cdot i_2) \mod n_v$. In this task, the output values derive from the same attribute transform applied in the \textit{lang-entangled} language (Eq.~\ref{eq:rotate}). This language-task pair mirrors the \textit{lang-identity}/\textit{task-identity} pair: each symbol encodes one \emph{output} value.

\paragraph{Architecture and hyperparameters}
Receiver is implemented as an LSTM~\cite{Hochreiter1997} or a GRU cell~\cite{Cho2014}. Its output layer specifies two categorical distributions over $n_v$ values, encoding two output values. As a loss, we use the sum of per-output negative log-likelihoods. We used the following hyperparameters: $n_v = 31$; hidden layer size 100; embedding size 50; batch size 32; 500 epochs training with Adam (learning rate $10^{-2}$). Each configuration was run 20 times with different random seeds. A random $1/5$ of the data is used as test set.

\subsection{Coordinates experiment}
\label{sArch}



\paragraph{Input} We sample points uniformly from the unit circle, centered at the origin: $\vi \in \mathbb{R}^2, ~~ \vi^T\vi \le 1$. We sample $10^3$ points for training, $10^3$ for testing.

\paragraph{Languages} We consider two languages with utterances of length two. In the first language, \textit{lang-coordinate}, Sender sequentially transmits both coordinates of a point: $m_j \gets i_j$. More precisely, the symbols refer to discretized coordinates from a $n_v \times n_v$ square grid, covering $[-1, 1] \times [-1, 1]$. This language is na\"ively compositional w.r.t.\ the coordinate-wise representation of the inputs.

We construct the second language, \textit{lang-rotated}, in the following way. We start with \textit{lang-coordinate}, but apply a rotation of the plane by $\pi/4$ before feeding a point into Sender.\footnote{Rotating by any angle $(0, \pi/2)$ makes the language non-compositional; $\pi / 4$ maximally entangles it.} Effectively, this makes Sender ``use'' a rotated coordinate grid for encoding the coordinates. As a result of the rotation, \textit{lang-rotated} ceases to be na\"ively compositional in the original (non-rotated) world. Each symbol of \textit{lang-rotated} carries equal amounts of information about both coordinates of $\vi$.

\paragraph{Task} Receiver has to recover the original (non-rotated) coordinates $\vi$ of a point.

\paragraph{Architecture and loss}
Receiver is an LSTM with hidden size 100 and embedding size 50; $n_v$ is 100; batch size is 32; we use Adam with learning rate $10^{-3}$. As a loss, we use MSE. We run each configuration with 10 random seeds.


\begin{figure}
  \centering
  \centering
  \includegraphics[width=1.0\columnwidth]{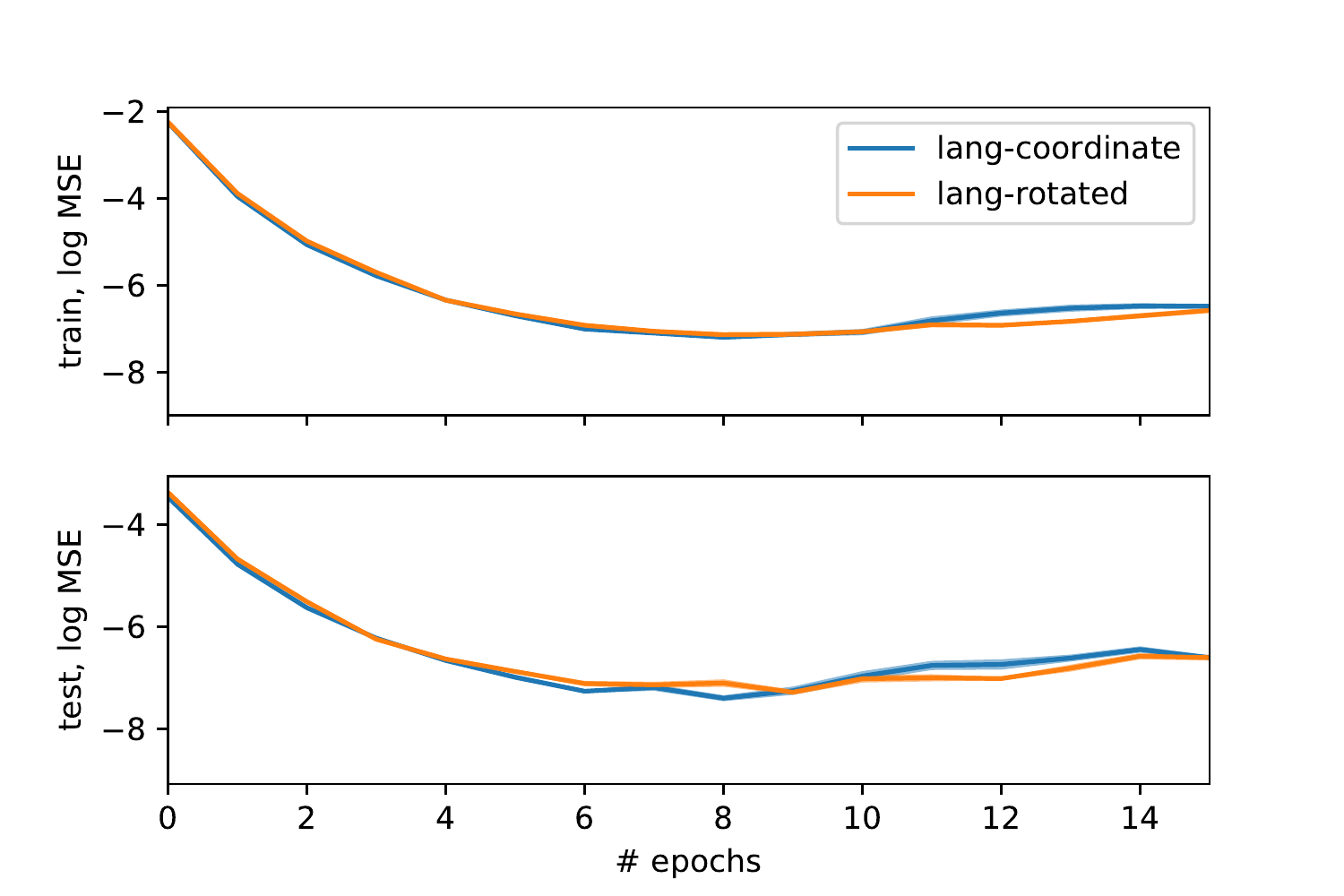}
    \caption{Coordinates experiment: $\log$ MSE vs.\ training epoch. \vspace{-2mm}}
\label{fig:continuous}
\end{figure}

  

\section{Results}
\paragraph{Attval experiment}
In Table~\ref{tab:attrval} we provide the results of the \textit{attrval} experiment, depending on language, task, and Receiver architecture. We report the number of epochs to achieve perfect accuracy on training set (top) and the accuracy on the hold-out set after training (bottom).

Consider the convergence speed first. For both Receiver architectures \textit{lang-identity} converges considerably faster than \textit{lang-entangled}. This agrees with the findings of~\citet{Li2019}. However, in \textit{task-linear} both languages demonstrate roughly the same convergence speed (the difference is not stat.\ sig.). In \textit{task-entangled}, \textit{lang-entangled} becomes more efficient to acquire than the na\"ively compositional \textit{lang-identity}. Interestingly, the acquisition times of \textit{task-identity}/\textit{lang-identity} and \textit{task-entangled}/\textit{lang-entangled} are symmetrical.

Next, consider the test accuracy of the same runs as above, measuring generalization to new attval combinations. We observe the same patterns: \textit{task-linear} is equally hard to generalize from both languages; \textit{lang-identity} reaches high test accuracy in \textit{task-identity}, while \textit{lang-entangled} leads to equally high accuracy on \textit{task-entangled}. In contrast, \textit{lang-identity} performs very poorly on \textit{task-entangled}, just as \textit{lang-entangled} does on \textit{task-identity}.

\paragraph{Coordinates experiment}
Figure~\ref{fig:continuous} reports learning curves for train and test sets (cueing acquisition speed and generalization, respectively). There is little difference between the compositional and non-compositional languages, in either training or held-out loss trajectories.

\section{Discussion and Conclusion}
Our toy experiments with hand-coded languages make the possibly obvious but currently overlooked point that, in isolation from the target task, there is nothing special about a language being (na\"ively) compositional.  A non-compositional language can be equally or even faster to acquire than a compositional one, and it can provide the same or better generalization capabilities.  Thus, if our goal is to let compositional languages emerge, we should be very clear about which characteristics of our setup should lead to its emergence. 

Our concern is illustrated by the recent findings of~\citet{Chaabouni2020}, who observed that the degree of compositionality of emergent languages is not correlated with the generalization capabilities of the agents that rely on them to solve a task. Indeed, lacking any specific pressure towards developing a (na\"ively) compositional language, their agents were perfectly capable of developing generalizable but non-compositional communication systems.

A stronger conclusion is that perhaps we should altogether forget about compositionality as an end goal. The current emphasis on it might just be a misguided effect of our human-centric bias. We should instead directly concentrate on the properties we want agent languages to have, such as fast learning, transmission and generalization.

\FloatBarrier
\bibliography{bib,marco}
\bibliographystyle{acl_natbib}

\end{document}

%% file: math_commands.tex
\usepackage{amsmath,amsfonts,bm}

\def\eqref#1{equation~\ref{#1}}

\def\1{\bm{1}}

\def\vb{{\bm{b}}}

\def\vi{{\bm{i}}}

\def\vm{{\bm{m}}}

\def\vo{{\bm{o}}}

\DeclareMathAlphabet{\mathsfit}{\encodingdefault}{\sfdefault}{m}{sl}
\SetMathAlphabet{\mathsfit}{bold}{\encodingdefault}{\sfdefault}{bx}{n}

